\documentclass{article}\usepackage{fullpage}
\pdfoutput=1

\usepackage{url}
\usepackage{amsmath}

\newcommand{\Description}[1]{}%
\newenvironment{acks}{\section*{Acknowlegdments}}{}%
\newcommand{\grantsponsor}[3]{#2}%
\newcommand{\grantnum}[2]{#2}%

\usepackage{natbib}
 \makeatletter
 \newcommand{\bibstyle@acmnumeric}{%
   \setcitestyle{%
     numbers,sort&compress,%
     open={[},close={]},citesep={,},%
     notesep={, }}}%
\makeatother
\usepackage{fnpct}
\setfnpct{add-punct-marks=!?:;} %

\usepackage{ellipsis}

\usepackage{tikz}\usetikzlibrary{shapes,calc,arrows,fit,positioning,decorations.pathmorphing,snakes,intersections,shapes.geometric,trees,cd}

\usepackage{subcaption}

\usepackage{booktabs} %

\tikzset{hoook/.style={}} %
\tikzset{hooop/.style={}} %

\usepackage[labelfont=bf,font=small]{caption}

\usepackage{xparse}

\NewDocumentCommand{\qnote}{om}{%
\IfNoValueTF{#1}%
    {\footnote{``#2''}}%
    {\footnote{``#2'' (#1)}}}

\title{What is Proxy Discrimination?}
\author{Michael Carl Tschantz\\ International Computer Science Institute\\ \url{mct@icsi.berkeley.edu}}

\begin{document}

\maketitle

\begin{abstract}\noindent
The near universal condemnation of proxy discrimination hides a disagreement over what it is.  This work surveys various notions of \emph{proxy} and \emph{proxy discrimination} found in prior work and represents them in a common framework.  These notions variously turn on statistical dependencies, causal effects, and intentions.  It discusses the limitations and uses of each notation and of the concept as a whole.
\end{abstract}

A version of this work will appear as
\begin{quote}
Michael Carl Tschantz. 2022. What is Proxy Discrimination? In \emph{2022 ACM Conference on Fairness, Accountability, and Transparency (FAccT ’22), June 21–24, 2022, Seoul, Republic of Korea}. ACM, New York, NY, USA.  \url{https://doi.org/10.1145/3531146.3533242}
\end{quote}

\section{Introduction}
\label{sec:intro}

Discrimination can occur when a classifier or model uses 
a sensitive attribute, such as a person's
gender or race, to reach a decision that should not depend upon that
attribute.
Perhaps the simplest attempt to prevent such discrimination, called \emph{blindness}~\cite{dwork12itcs} or \emph{treatment parity}~\cite{zafar17neurips}, is to
exclude such attributes from the data available to the classifier.
This approach finds little favor in the research community, as
simply hiding an attribute does not prevent machine learning from effectively recreating it from 
the other provided attributes~\cite[e.g.,][]{pedreshi08kdd,dwork12itcs,hardt14medium,barocas16cal-l-rev}.

This concern is often described using the framing of \emph{proxies}, including in a White House report on big data~\cite[p.\,53]{bigdata14whitehouse}\qnote{Just as neighborhoods can serve as a proxy for racial or ethnic identity, there are new worries that big data technologies could be used to ``digitally redline'' unwanted groups, either as customers, employees, tenants, or recipients of credit.}, and \emph{proxy discrimination}.
(The footnotes contain only quotations from cited sources.) 
For example, an algorithm might use someone's music interests as a \emph{proxy}
for race.  In more complex cases, an algorithm might use a combination
of features, each of which look innocent on its own, to construct a
\emph{proxy}.
Either way, the algorithm might engage in \emph{proxy discrimination} against a race.
But, what exactly is a \emph{proxy} or \emph{proxy discrimination}?

In this work, I introduce a common framework in which to consider and compare various notions of \emph{proxy} and \emph{proxy discrimination}.
My framework uses causal reasoning, including in a second-order fashion that I have not seen elsewhere in fairness research.
I see this paper as the starting point for a rigorous discussion of what is meant by \emph{proxy}, \emph{proxy discrimination}, \emph{wrongful proxy discrimination}, and \emph{illegal proxy discrimination}.
However, herein, I do not focus on moral or legal questions, but rather the empirical characteristics of proxies and proxy discrimination.
I focus on what can be described using standard scientific methods: associations, causes, and effects.
I also touch upon intentions.
I do not tackle normative issues better left to domain experts, democratic processes, or public opinion, such as which attributes should not be proxied.

I first provide an overview of the various notions I consider (\S\ref{sec:overview}) and the related work (\S\ref{sec:prior-work}).
I then get more technical, introducing some needed formalism (\S\ref{sec:background}) before applying it to take a closer look at the notions, one by one (\S\S\ref{sec:capacity}--\ref{sec:substituting}).
Along the way, I comment on some strengths and weaknesses of various definitions.
Next, I again consider the notions together, but this time through a normative lens (\S\ref{sec:norms}).
While I do not conclude that any one definition is the correct one, I find some more apt than others at matching my moral intuitions.
I conclude that there is no one clear definition of \emph{proxy discrimination}, but that the term identifies a cluster of related concerns, each warranting attention (\S\ref{sec:conclusion}).

\section{Overview of Notions of Proxy and Proxy Discrimination}
\label{sec:overview}

While the notions of proxy and proxy discrimination I consider can be flexibly applied to a range of tasks, I am primarily concerned with a classifier using a model, perhaps created by machine learning, to produce estimations or predictions $\hat{Y}$ of some attribute $Y$ of individuals.
The classifier has access to other attributes $\vec{X}$ of the individuals to produce $\hat{Y}$.

Typically, discussions of proxy discrimination start with a sensitive or prohibited attribute $T$ that should not be one of the used attributes in $\vec{X}$.
They note that some other attribute $P$ in $\vec{X}$ (or a combination $P$ of attributes in $\vec{X}$ used jointly) could be a proxy targeting $T$.
They may speak of $P$ recreating $T$, thereby undoing the exclusion of $T$ from $\vec{X}$.

For example, consider the sentence ``Zip code is a proxy for race in lending''.
Here, $P$ is zip code; $T$ is race; $\vec{X}$ might be attributes such as zip code, income, and repayment history; $Y$ might be loan default; and $\hat{Y}$ might be a credit score predicting $Y$ computed by a model using $\vec{X}$.
This sentence emphasizes the association or correlation between the proxy $P$ and target $T$.
Compare that to the sentence ``The credit scoring model uses zip code as a proxy for race'', which emphasizes how the model uses the proxy in the place of its target.
Despite having the same values for $P$, $T$, $\vec{X}$, $Y$, and $\hat{Y}$, the two sentences suggest different uses of the word ``proxy''.

By pondering sentences such as these and closely reading prior works using the terms \emph{proxy} and \emph{proxy discrimination}, I identified various definitions of these terms.
I show how these definitions break apart into recurring elements.
Table~\ref{tbl:elements} shows these elements as relationships that may hold between $P$, $T$, and $\hat{Y}$.
Some elements are more relevant to a finding of proxy discrimination than to a finding of having a proxy, but the more elements that hold in a scenario, the more apt it becomes to describe the scenario as involving a proxy or proxy discrimination.
I separate the elements into two tables based on whether it's a first-order relationship between two variables or a second-order relationship between two relationships.
I start with the first order ones.

\begin{table*}
\caption{Elements of proxy and proxy discrimination: relationships suggesting that $P$ is a proxy targeting $T$ for the estimator $\hat{Y}$.  
The table lists them, provides names, and separates them into first-order and second-order relationships.  Dashed arcs denote association-like relationships (e.g., statistical dependence, correlation).  Solid arrows denote inducement (e.g., causes, motivates, intentionally brings about).  Slashed-out gray arrows denote a lack of inducement.}
\label{tbl:elements}
\Description{The table shows that first-order capacity (Section 5) has an association-like relationship between P and T, that first-order proxy use (Section 6) has an inducement from P to hat Y, that first-order target unused (Section 7) lacks an inducement from T to hat Y, that first-order bottom line impact (Section 8) has an association-like relationship between T and hat Y, that second-order consequential (Section 9) has an inducement from P inducing hat Y to an association-like relationship between T and hat Y.  It then shows that second-order proxy substituting for target (Section 11) involves an inducement from the lack of T inducing hat Y to P inducing hat Y.  It finally shows that second-order capacity induced proxy use (Section 10) involves an inducement from an association-like relationship between P and T to P inducing hat Y.}
\centering
\renewcommand{\arraystretch}{1.2} %
 \begin{tabular}{@{}llc@{}}
 \toprule
  Order & Name & Relationship\\
  \midrule
  First-order
  & Capacity (\S\ref{sec:capacity})
  &
   \begin{tikzcd}[math mode=true, row sep=2em, column sep=5em]
      |[alias=P]| P & |[alias=T]| T
    \arrow[from=P, to=T, dashed, no head, bend left=5]
   \end{tikzcd}
   \\
  & Proxy Use (\S\ref{sec:use})
  &
   \begin{tikzcd}[math mode=true, row sep=2em, column sep=5em]
       |[alias=P]| P & |[alias=hY]| \hat{Y}
     \arrow[from=P, to=hY, hoook]
   \end{tikzcd}
   \\
  & Target Unused (\S\ref{sec:target-unused})
  &
\begin{tikzcd}[math mode=true, row sep=2em, column sep=5em]
       |[alias=T]| T & |[alias=hY]| \hat{Y}
     \arrow[from=T, to=hY, hoook, "/"{anchor=center,sloped}, gray]
   \end{tikzcd}
   \\
  & Bottom Line Impact (\S\ref{sec:impact})
  &
   \begin{tikzcd}[math mode=true, row sep=2em, column sep=5em]
      |[alias=T]| T
    & |[alias=hY]| \hat{Y}
    \arrow[from=T, to=hY, dashed, no head, bend left=5]
   \end{tikzcd}
   \\
  \midrule
  Second-order
  & Consequential (\S\ref{sec:consequential})
  &
   \begin{tikzcd}[math mode=true, row sep=1em, column sep=2em]
       |[alias=P]| P &                         &  |[alias=T]| T \\
                     &  |[alias=hatY]| \hat{Y} &     
      \arrow[from=T, to=hatY, dashed, no head, bend left=5, ""'{name=bottomline}]
      \arrow[from=P, to=hatY, hoook, ""{name=proxyuse}]
      \arrow[from=proxyuse, to=bottomline, hoook, ""']
   \end{tikzcd}
   \\
  & Proxy Substituting for Target (\S\ref{sec:substituting})
  & 
   \begin{tikzcd}[math mode=true, row sep=1.5em, column sep=2em]
      |[alias=P]|    P &                      &  |[alias=T]| T \\
                       & |[alias=hY]| \hat{Y} & 
    \arrow[from=P, to=hY, hoook, ""{name=use}]
    \arrow[from=T, to=hY, hooop, "/"{anchor=center,sloped}, gray, ""'{name=prohibition}]
    \arrow[from=prohibition, to=use, hooop]
   \end{tikzcd}
   \\
  & Capacity Induced Proxy Use (\S\ref{sec:mot})
  & 
   \begin{tikzcd}[math mode=true, row sep=1.5em, column sep=2em]
      |[alias=P]|    P &                      &  |[alias=T]| T \\
                       & |[alias=hY]| \hat{Y} & 
    \arrow[from=P, to=T,  dashed, no head, bend left=5, ""'{name=predictiveValue}]
    \arrow[from=P, to=hY, hoook, ""{name=use}]
    \arrow[from=predictiveValue, to=use, hooop]
   \end{tikzcd}
   \\
 \bottomrule
 \end{tabular}
\end{table*}

The first, and most central, element I consider is the \emph{capacity} of $P$ to target $T$, that is, how accurately $P$ can recreate $T$.
Different authors make this relationship precise using various relationships like statistical association, as I discuss in Section~\ref{sec:capacity}.
In general, the stronger the association, the more reason there is to call $P$ a \emph{proxy} for $T$ on these grounds.
With this in mind, I do not see the capacity relationship or being a proxy to be crisply true or false, but rather as matters of degree.
Without any capacity, I do not find the use of the term \emph{proxy} to be appropriate, but as I discuss in Section~\ref{sec:capacity}, there are so many ways to achieve a degree of capacity that this is hardly a restriction.

The \emph{proxy use} element of proxy discrimination is more causal in nature.
It holds when the value of $P$ helps to induce the value that $\hat{Y}$ takes on.
This inducement could be that $P$ is a cause of $\hat{Y}$'s value, that it motivates the value, that an agent intends to set the value with $P$ in mind, or some other productive relationship.
I discuss various options in Section~\ref{sec:use}. 
This turns us from looking at whether $P$ \emph{could} be used to make predictions to whether it really \emph{was} used in this manner, and may account for some authors saying a feature ``is'' a proxy and not merely that it ``can be'' one.
Generally speaking, such proxy use is required to go from ``proxy'' to ``proxy discrimination''.
As with associations, the degree to which the inducement justifies using the terms ``proxy (discrimination)'' depends upon its nature and strength.

The remaining elements are not necessary for proxy existence or discrimination, but each further justifies using the terms.
The \emph{target unused} relationship holds when the target isn't used to induce values for $\hat{Y}$.
When $\hat{Y}$ can use $T$ directly, it's less clear that $P$ should be considered a proxy for $T$ even if it has the capacity to be one.

The element of \emph{bottom line impact} looks at whether the resulting decisions $\hat{Y}$ end up associated, in some sense, with the target.
This measure of the so-called ``bottom line'' can be viewed as a measure of the proxy's impact or harm.
Intuitively, the stronger the association-like property, the more harm done.

The remaining elements of proxy discrimination are each about whether one of the aforementioned relationships induces another one of them, which I view as a second-order relationship.
They cannot be decomposed into a mere conjunction of the underlying first-order relationships since they capture something about the interaction of the two relationships not present in either alone.
For example, it is possible to have proxy use and a bottom line impact that do not have anything to do with one another since the impact could come from some other aspect of the model.
Only when the proxy use induces a bottom line impact is the proxy \emph{consequential}.

\emph{Substituting} holds when the target going unused induces the classifier to use the proxy.
Closely related is \emph{capacity induced use}.
These two elements can be teased apart with two examples.

As an example of substituting without capacity induced use, consider a case where $P$ and $T$ are both statistically associated with $Y$ but in different ways so that they are not associated with one another.
Further suppose that $\hat{Y}$ only needs to cross some threshold for accuracy that can be reached using either of $P$ and $T$ but that using $P$ is more expensive than $T$.
In this case, the classifier will use $T$ without $P$ if allowed, and will use $P$ only if using $T$ is disallowed.

As an example of capacity induced use without substituting, consider a case where $P$ is correlated with $T$ and $T$ with $Y$, and the classifier uses $P$ for that reason.
Further suppose that the classifier's model is produced by a machine learning algorithm that attempts to use every allowed feature correlated with $Y$ 
(e.g., one using an L2 norm, such as ridge regression).
It does not have substituting since it would use both $P$ and $T$ if allowed, meaning it uses $P$ regardless of $T$'s use.

\begin{figure*}
 \begin{subfigure}{.3\linewidth}\centering
\begin{tikzcd}[math mode=true, row sep=2em, column sep=8em]
      |[alias=P]| P & |[alias=T]| T
    \arrow[from=P, to=T, dashed, no head, bend left=5, "\text{correlation (\S\ref{sec:correlation})}"]
   \end{tikzcd}    \caption{Kirkpatrick's ``proxy''~\cite[p.\,16]{kirkpatrick16cacm}: capacity}
   \label{fig:ov-kirkpatrick-proxy}
 \end{subfigure}
\hfill
 \begin{subfigure}{.3\linewidth}\centering
   \begin{tikzcd}[math mode=true, row sep=2em, column sep=10em]
      |[alias=P]| P & |[alias=T]| T
    \arrow[from=P, to=T, dashed, no head, bend left=5, "\text{two-sided correlation (\S\ref{sec:two-sided})}"]
   \end{tikzcd}
    \caption{Datta et al.'s ``proxy''~\cite[Def.\,2]{datta17arxiv}: capacity}
   \label{fig:ov-datta-proxy}
 \end{subfigure}
\hfill
 \begin{subfigure}{.3\linewidth}\centering
   \begin{tikzcd}[math mode=true, row sep=1.5em, column sep=4em]
      |[alias=P]|    P &                      &  |[alias=T]| T \\
                       & |[alias=hY]| \hat{Y} & 
    \arrow[from=P, to=T,  dashed, no head, bend left=5, "\text{two-sided cor.\@ (\S\ref{sec:two-sided})}"{name=predictiveValue}]
    \arrow[from=P, to=hY, "\text{influence (\S\ref{sec:influence})}"{name=use, near end}]
   \end{tikzcd}
    \caption{Datta et al.'s ``proxy use''~\cite[Def.\,6]{datta17arxiv}: capacity and proxy use}
   \label{fig:ov-datta-proxy-use}
 \end{subfigure}
 
 \begin{subfigure}{.3\linewidth}\centering
   \begin{tikzcd}[math mode=true, row sep=1em, column sep=2em]
       |[alias=P]| P &                         &  |[alias=T]| T \\
                     &  |[alias=hY]| \hat{Y} &     
      \arrow[from=P, to=T,  dashed, no head, bend left=5, ""'{name=predictiveValue}]
      \arrow[from=T, to=hY, dashed, no head, bend left=5, ""'{name=bottomline}]
      \arrow[from=P, to=hY, hoook, ""{name=proxyuse}]
   \end{tikzcd}
    \caption{EEOC policy on pursuing disparate impact~\cite{eeoc78cfr} (\S\ref{sec:impact}): capacity, use, and impact}
   \label{fig:ov-eeoc}
 \end{subfigure}
\hfill
 \begin{subfigure}{.3\linewidth}\centering
   \begin{tikzcd}[math mode=true, row sep=2em, column sep=10em]
      |[alias=P]| P & |[alias=T]| T
    \arrow[from=T, to=P, dashed, bend right=5, "\text{causal descendant (\S\ref{sec:causal-paths})}"']
   \end{tikzcd}

    \caption{Kilbertus et al.'s proxy~\cite[p.\,4]{kilbertus17neurips}: capacity}
   \label{fig:ov-kilbertus-proxy}
 \end{subfigure}
\hfill
 \begin{subfigure}{.3\linewidth}\centering
   \begin{tikzcd}[math mode=true, row sep=1.5em, column sep=4em]
      |[alias=P]|    P &                      &  |[alias=T]| T \\
                       & |[alias=hY]| \hat{Y} & 
    \arrow[from=P, to=hY, dashed, "\text{causal descendant (\S\ref{sec:use-causal-descendants})}"{near end}]
    \arrow[from=T, to=P, dashed, bend right=5, "\text{causal descendant (\S\ref{sec:causal-paths})}"']
   \end{tikzcd}
    \caption{Kilbertus et al.'s ``potential proxy discrimination''~\cite[Def.\,2]{kilbertus17neurips}: capacity and use}
   \label{fig:ov-kilbertus-pot-proxy-discrimination}
 \end{subfigure}

 \begin{subfigure}{.3\linewidth}\centering
   \begin{tikzcd}[math mode=true, row sep=2em, column sep=4em]
      |[alias=P]|    P &                      &  |[alias=T]| T \\
                       & |[alias=hY]| \hat{Y} & 
    \arrow[from=P, to=hY, "\text{total effect (\S\ref{sec:total-effects})}"{name=use, near end}]
    \arrow[from=T, to=P, dashed, bend right=5, "\text{causal descendant (\S\ref{sec:causal-paths})}"']
   \end{tikzcd}
    \caption{Kilbertus et al.'s ``proxy discrimination''~\cite[Def.\,3]{kilbertus17neurips}: capacity and use}
   \label{fig:ov-kilbertus-proxy-discrimination}
 \end{subfigure}
\hfill
 \begin{subfigure}{.3\linewidth}\centering
   \begin{tikzcd}[math mode=true, row sep=1.5em, column sep=2em]
      |[alias=P]|    P &                      &  |[alias=T]| T \\
                       & |[alias=hY]| \hat{Y} & 
    \arrow[from=P, to=T,  dashed, no head, bend left=5, ""'{name=predictiveValue}]
    \arrow[from=P, to=hY, hoook, ""{name=use}]
    \arrow[from=predictiveValue, to=use, hooop]
   \end{tikzcd}
  \caption{Prince and Schwarcz's ``proxy discrimination''~\cite[p.\,1261]{prince20iowa-lr}: capacity induced use}
 \label{fig:ov-prince-schwarcz}
 \end{subfigure}
\hfill
 \begin{subfigure}{.3\linewidth}\centering
 \begin{tikzcd}[math mode=true, row sep=3em, column sep=4em]
     |[alias=P]| P &                         &  |[alias=T]| T \\
                   &  |[alias=hatY]| \hat{Y} &     
    \arrow[from=P, to=T, dashed, no head, bend left=15, ""'{name=predictiveValue}]
    \arrow[from=P, to=hatY, hoook, ""{name=proxyuse}]
    \arrow[from=T, to=hatY, "/"{anchor=center,sloped}, gray, shift right, ""'{name=Tnotused}]
    \arrow[from=T, to=hatY, dashed, no head, bend left=20, shift left, ""'{name=bottomline}]
    \arrow[from=Tnotused, to=proxyuse, hoook, start anchor={[yshift=2ex, xshift=3ex]}, end anchor={[yshift=1ex, xshift=-1.5ex]}, ""']
    \arrow[from=predictiveValue, to=proxyuse, hoook, end anchor={[yshift=2ex, xshift=-3ex]}, ""']
    \arrow[from=proxyuse, to=bottomline, hoook, crossing over, crossing over clearance=1ex, shift right, ""']
 \end{tikzcd}
  \caption{All of them at once}
 \label{fig:ov-all}
 \end{subfigure}
\caption{Combinations of elemental relationships.  When a prior work is clear about the sort of association or inducement used, I note it here, but defer a discussion until later.}
\label{fig:overview}
\end{figure*}
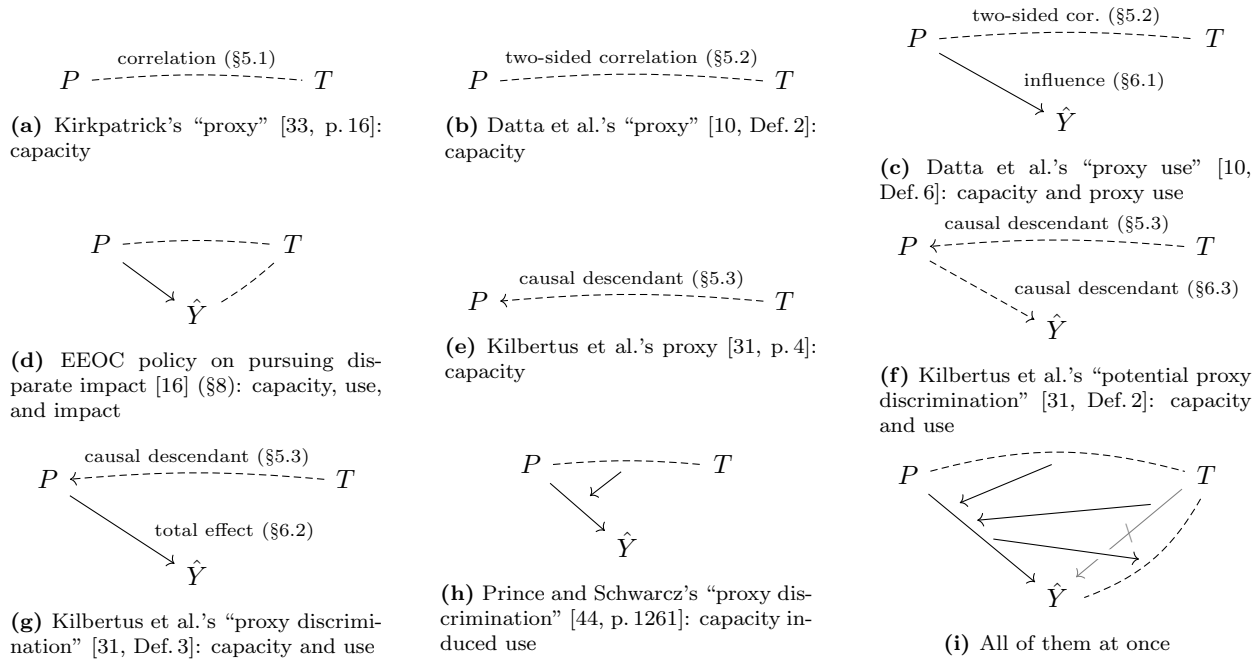

Figure~\ref{fig:overview} provides an overview of definitions of \emph{proxy} and \emph{proxy discrimination} built out of these elements of proxy discrimination.
Some use just one element.
For example, some authors view the capacity relationship as sufficient to make $P$ a proxy for $T$, although for different notions of association (Figs.~\ref{fig:ov-kirkpatrick-proxy} \&~\ref{fig:ov-datta-proxy}).
Prince and Schwarcz define \emph{proxy discrimination} to happen when a proxy's use is capacity induced~\cite{prince20iowa-lr}, which uses a single second-order element, thereby also touching upon two component first-order elements (Fig.~\ref{fig:ov-prince-schwarcz}).
Other definitions are constructed using conjunctions of these elements.
For example, Fig.~\ref{fig:ov-eeoc} shows what one gets if one views disparate impact through the lens of proxy discrimination and checks it against the EEOC's policy for when it will bring a case (which is more strict than disparate impact in general; see \S\ref{sec:impact}).
Figure~\ref{fig:ov-all} shows a combination of all the elemental relationships to get what might be considered the quintessential case of proxy discrimination.

\section{Related Work}
\label{sec:prior-work}

I believe this paper to be the first to rigorously compare notions of proxy discrimination proposed by various authors.
Forthcoming work by Boyarskaya et al.\@ has similar goals but a different scope~\cite{boyarskaya22unpub}.
Being a work in progress, I don't consider it outside of this section.
Many of the papers cited as a source of a notion of proxy discrimination are primarily concerned with a task other than exploring the notion's semantics, such as exploring algorithmic discrimination in general~\cite[e.g.,][]{barocas16cal-l-rev,kim17wm-mary-lr}, studying disparate impact~\cite[e.g.,][]{feldman15kdd}, or developing approaches to detect or eliminate a notion of proxy discrimination taken as a given~\cite[e.g.,][]{pope11am-econ-j,datta17arxiv,kilbertus17neurips}. 
I introduce these works as I go along.
Prince and Schwarcz~\cite{prince20iowa-lr} takes definitional issues rather seriously, but focuses on just its preferred definition and lacks a mathematical treatment, but does offer more legal analysis than found herein.

Hellman explores the topic from the opposite direction of almost every other work cited: rather than being concerned with seemingly innocent variables being used as a proxy for a protected one, she is concerned with using the protected one as a proxy for unprotected attributes~\cite{hellman98CaLR}.
While the math remains the same, the intuitions are often flipped.
Boyarskaya et al.\@ discusses this flip in more detail~\cite{boyarskaya22unpub}.

The only prior work on second-order causation that I could find aims to make precise functional explanations found in evolutionary biology~\cite{hitchcock96erkenntnis}.
It considers a causation causing an attribute whereas herein we look at a causation causing a correlation and at a lack of causation causing a different causation.
Nevertheless, the approach taken therein appears adaptable to our needs here, which I leave as future work.

\section{Background on Machine Learning and Causal Modeling}
\label{sec:background}

I will limit our attention to machine learning applied to people using the simplest form of supervised learning.
To be concrete, I will introduce the needed notation assuming that the user of the machine learning is attempting to select convicts for parole who will not reoffend.

Consider an actuary working for a parole board, which collectively acts as the decisionmaker.
She may start by collecting as much data as her budget allows about prior parolees, such as their ages, sexes, educations, rearrests, and subsequent convictions.
Notably, the data will not include which prior parolees reoffended since not all reoffenders will get caught and false convictions do occur.
While this could be called ``proxying'', and raises concerns about fairness~\cite[e.g.,][]{friedler21cacm}, it is not the sort of proxying that concerns this paper.
Thus, I will just assume that she uses whether the parolee was reconvicted within five years and that everyone agrees that it is reasonable to predict this instead of reoffense.

To use ML nomenclature, the decisionmaker's goal is to \emph{classify} prisoners into those who will be reconvicted and those who will not.
Each prior parolee comes \emph{labeled} with whether or not they have been reconvicted. 
Each of them also comes with a list of values for their \emph{features}, all the other recorded attributes that may be useful in predicting reconvictions.
To be so useful, it must be possible to determine the values of these features before the parole decision is made, unlike the class label, which is only available five years later.

Mathematically, I can represent the class as a random variable $Y$ over the set prisoners.
A vector $\vec{X}$ of random variables can represent the features.
The data about prior parolees is a set of pairs $\langle \vec{x}_i, y_i\rangle$ that record the values of $\vec{X}$ and $Y$ for each prior parolee $i$.
A machine learning algorithm $a$ consumes this set as \emph{training data}.
It attempts to find patterns in it that allows it to predict a potential parolee's class from their features.
The algorithm produces a \emph{model} $m$ that acts as a \emph{classifier}.
The model $m$ takes the features $\vec{X}$ of a prisoner and predicts whether they will be reconvicted within five years.
I will assume that this prediction comes as just a binary \emph{yes} or \emph{no}, but it could come with a confidence score or other information.
I represent this prediction as $\hat{Y}$.
The random variable $\hat{Y}$ \emph{estimates} $Y$.
The goal of the algorithm $a$ is to select an $m$ such that its prediction $\hat{Y}$ for each prisoner is equal to the prisoner's actual class $Y$.
Ideally, $\Pr[\hat{Y} = Y] = 1$.

Let $T$ be a sensitive attribute that should not be used by the model to make predictions, such as the prisoner's race.
In an employment setting, it might be an applicant's sex.
In a health insurance setting, it might be a pre-existing condition.
$T$ should not be in $\vec{X}$, which could be checked by examining what data is collected and used.
(Under a more complex form of ML, \emph{disparate learning processes}, the algorithm $a$ gets not just the features $\vec{X}$ and labels $Y$, but also a second set of features that the learned model $m$ does not get to use when making predictions, with the idea that it may be acceptable to train on the sensitive attribute $T$ even if it is unacceptable to classify with it~\cite[e.g.,][]{lipton18nips}.  I set this aside.)

The intuition behind proxy discrimination is that one or more features in $\vec{X}$ might be used to recreate $T$.
For example, a prison's pre-incarceration zip code could be used as a proxy $P$ that targets the protected attribute of race $T$.
I will explore various notions of what it means to use $P$ as a proxy targeting $T$.

Many of notions of proxy (discrimination) depend upon causal reasoning about interventions, which I must distinguish from probabilistic conditioning.
For example, I use $\Pr[Y{=}0 \mathrel{|} \hat{Y}{=}0]$ to denote the conditional probability that $Y$ takes on the value $0$ (not reconvicted within five years) given that $\hat{Y}$ is equal to $0$.
This probability is a measure of the accuracy of $m$, looking at how often prisoners who are predicted to not be reconvicted, and therefore are released, are to actually not be reconvicted.
I use $\Pr[Y{=}0 \mathrel{|} \mathrm{do}(\hat{Y}{=}0)]$ to instead denote the probability that $Y$ takes on the value $0$ if I artificially intervene upon the behavior of $m$ to assign them all to have $\hat{Y}$ set to $0$.
This probability is no longer measuring the accuracy of $m$, or even using $m$ since $m$ is replaced by a constant function always reporting the value $0$.
It is instead looking at what would happen if I ran an experiment and released everyone.
It would be very difficult to calculate this probability without running such an experiment, but if I could approximate it with a causal model of criminality, it would be useful for understanding prison reforms and as a baseline to compare the accuracy of $m$ to.

Much easier would be computing $\Pr[\hat{Y}{=}y \mathrel{|} \mathrm{do}(\vec{X}{=}\vec{x})]$, which is the probability of $m$ producing the output $y$ given that all its inputs are $\vec{x}$.
I could run $m$ numerous times on the inputs $\vec{x}$ to approximate this probability or analyze the internals of $m$ to determine it.
Since $\vec{X}$ are all the inputs to $m$, this is a special case where conditioning and causal interventions collapse and $\Pr[\hat{Y}{=}y \mathrel{|} \mathrm{do}(\vec{X}{=}\vec{x})] = \Pr[\hat{Y}{=}y \mathrel{|} \vec{X}{=}\vec{x}]$~\cite[Property~1, p.\,24]{pearl09book}.
If I instead intervene upon or condition upon anything less than all of $m$'s inputs, this property will not necessarily hold.
For example, in general,
$\Pr[\hat{Y}{=}y \mathrel{|} \mathrm{do}(P{=}p)] \neq \Pr[\hat{Y}{=}y \mathrel{|} P{=}p]$.
Intuitively, this inequality is because conditioning upon $P$ provides some information about any of the other features in $\vec{X}$ that are associated with $P$ whereas intervening upon $P$ breaks these associations, cutting off this source of additional information.
Conditioning upon all the inputs makes this source of additional information no longer helpful since the values of all the inputs are already known.

Structural equation models (SEM) can make the above intuitions precise~\cite{pearl09book}.
For our purposes, it suffices to think of them as a series of assignment statements over random variables.
For example, an SEM might include the assignment that $\hat{Y} := m(\vec{X})$.
Unlike an equality, this assignment is directional, saying that interventions on the inputs of $m$ can affect the output but not the other way around.
It is common to represent such assignments with arrows, such as $\vec{X} \rightarrow \hat{Y}$ to graphically represent how variables flow from one to the next, similar to how I did so in Figure~\ref{fig:overview}.
The variables $\vec{X}$ shown in the assignment or graphical model are called the \emph{parents} of $\hat{Y}$. 

Note, however, that from just the assignment statement or the graphical model, you cannot tell whether $\hat{Y}$'s parents $\vec{X}$ have an effect on the output $\hat{Y}$ since $m$ could ignore its inputs $\vec{X}$ and be a constant function.
You have to also know the value of $m$.
Nevertheless, you can tell that, under the SEM, $\hat{Y}$ does not directly depend upon variables other than its parents.
(The SEM could be wrong about that.)
$\hat{Y}$ might indirectly depend upon its parents' parents and so forth.
Any variable that has a directed path leading to $\hat{Y}$ has $\hat{Y}$ as a \emph{causal descendant} and might affect it.

\section{Capacity}
\label{sec:capacity}

I consider three approaches to making the association-like relationship found in the capacity element of proxy discrimination precise.
I find these approaches lead to a multiplicity of options, enough that an auditor could go proxy fishing to accuse many features of having the capacity to proxy for another even if they are not very associated in general.
On the one hand, these features often can be used to discriminate, meaning I shouldn't dismiss them as unworthy of the title ``proxy''.
On the other hand, their ubiquity suggests that auditors need to prioritize by either focusing on those with the most capacity or those that have other elements of proxy discrimination.

\subsection{Dependence or Correlation}
\label{sec:correlation}

The simplest measure of $P$'s capacity to target $T$ is its statistical association, dependence, or correlation with $T$.
Some authors use ``proxy'' in a manner that suggests that any variable correlated with a target is a proxy for it.
For example, O'Neil appears to be using ``proxy'' in such a manner when stating
``[\ldots] OK, let's not use race, but should we use zip code, which of course is a proxy for race in our segregated society?''~\cite{oneil16hbr}.
Kirkpatrick also appears to \emph{proxy} in this sense~\cite[p.\,16]{kirkpatrick16cacm}.\qnote{It is not only the criminal justice system that is using such algorithmic assessments. Algorithms also are used to serve up job listings or credit offers that can be viewed as inadvertently biased, as they sometimes utilize end-user characteristics like household income and postal codes that can be proxies for race, given the correlation between ethnicity, household income, and geographic settling patterns.}

Despite its simplicity, there are many ways of measuring the statistical dependence or correlation between $P$ and $T$.
For example, if they are both continuous, then the Pearson product-moment correlation coefficient would be standard, unless there's outliers, which calls for the Spearman rank correlation coefficient~\cite[p.\,157]{khamis08jdms}.
More importantly, the size of the correlation between $P$ and $T$ depends upon the population it is computed over.
Suppose the correlation between a possible proxy $P$ and a target $T$ is absent in the world, strong in the U.S., absent in Georgia, but strong in Atlanta.
Suppose that a classifier for aiding hiring decisions uses $P$ and is marketed to the whole world, but is really only used in the U.S., particularly in Georgia, where it is most popular in Atlanta.
Does $P$ have enough capacity to count as a proxy?
While prior work has examined enforcing other fairness conditions on numerous subsets of a population~\cite{tramer17eurosp,kearns18icml,hebert-johnson18icml,dwork18fatml,dwork18arvix,kearns19facct}, it is unclear whether these methods can sensibly apply to this notion of proxy discrimination without leading to an explosion in the number of proxies.
Perhaps, we should speak of proxies for populations, but then we must decide which populations matter.

Furthermore, since the classifier also uses the features $\vec{X}'$ in $\vec{X}$ other than $P$,
we may wish to consider the dependence or correlation between $P$ and $T$ conditional on $\vec{X}'$.
This is similar to looking at subpopulations picked out by the covarients $\vec{X}'$.
If, as is commonly the case with machine learning, $\vec{X}'$ is rich, then it will pick out so many subpopulations that some may have a dependence between $P$ and $T$ by luck alone.
We can avoid this issue by looking at measures of $P$ and $T$'s correlation within a statistical model that operates across all of these subpopulations, as done in multiple linear regression, but this introduces another degree of freedom over the statistical model choice.

Changing the feature set $\vec{X}'$ can change whether $P$ has a capacity to target $T$.
For example, if $Y := T$ and $P := T \mathop{\mathrm{xor}} X_1$, then $P$ has the capacity to target $T$ iff the model can determine the value of $X_1$, by $X_1$ being included in $\vec{X}$ or inferred from it.
As a more realistic example, the language used at home becomes much more strongly associated with being an immigrant after conditioning upon the country in which the home is located.
One could define ``proxy'' so that language is a proxy for immigration on its own, or so that it is not but that language is a proxy for immigration given country or that language and country are jointly a proxy for immigration. 

Relatedly, a proxy may have capacity in general but not with respect to the specific ML algorithm used.
For example, suppose that $Y := T$ and $T := X_1 * X_2$ with $\vec{X} = \langle X_1, X_2\rangle$, where $X_1$ and $X_2$ are over the integers.
The proxy $P := X_1 * X_2$ is perfectly correlated with $T$, suggesting that $X_1$ and $X_2$ jointly have high capacity.
However, if the actual algorithm used is restricted to considering linear expressions over $\vec{X}$, it would struggle to find a capable proxy for $T$ since it could not express this non-linear relationship.
While in general more powerful algorithms will be more capable of finding and using proxies, it is not always the case that a more powerful algorithm is more likely to use any given variable as a proxy.
Consider extending the above example so that $\vec{X} = \langle X_1, X_2, X_3\rangle$ where $X_3 := X_1 * X_2 + \mathsf{Noise}$.
A linear algorithm may use $X_3$ as a noisy proxy for $T$ whereas a quadratic one could construct and use the more exact proxy $P := X_1 * X_2$.

Furthermore, we often do not have access to data about the whole population of interest, but rather just samples from it.
A sample could be biased, either introducing or eliding dependencies found in the whole population.
The sample used by the machine learning algorithm to train the model might have different dependencies than a sample used to audit the model.
Such differences can lead to disputes over whether a supposed proxy is real or merely an artifact of one's sample.

Relatedly, we may wish to prohibit the use of variables mistakenly believed to be
correlated with sensitive attributes since ``irrational proxy discrimination, based upon inaccurate stereotypes or generalizations, is morally troublesome because it imposes unnecessary social costs''~\cite{alexander92upennLR}.
In this case, we can calculate the correlation using
Bayesian probabilities modeling what someone believes
instead of 
frequentist probabilities modeling a real population, but this means that bigots get to think proxies into existence, and auditors may imagine even more into a status of \emph{possibly exists}.

Regardless of how these choices are resolved, a notion of proxy defined solely in terms of correlations is very general in that many variables will have at least some weak correlation with sensitive targets, such as race.
O'Neil is aware of this, going on to say ``And so once they acknowledge that zip code is just as good as race, then you're like, OK, so how do we choose our attributes? Because there are so many proxies to race''~\cite{oneil16hbr}.
Thus, we should not dismiss this notion of proxy as being unused
simply because it will hold between almost any two interesting random
variables.

We could attempt to pare such proxies back by requiring the association to be of a certain size.
However, this still seems to be a weak notion of proxy.
Suppose that the supposed target $T$ is race, the supposed proxy $P$ is the result of genetic test for the sickle hemoglobin (HbS) allele, and the predicted class $Y$ is whether the person will get sickle cell anemia.
Given the association between race and the HbS allele, $P$ will be a proxy for race.
However, given the clear use of $P$ in determining $Y$, it seems odd to declare $P$ a proxy for race, which has less to do with $Y$ or $P$ than either have to do with the other.
It seems even odder to conclude that this could be proxy discrimination, regardless of the strength of the association between $P$ and $T$.

Kraemer~et~al.\@ instead define \emph{proxy} to require
that $T$ is more strongly associated with $Y$ than $P$ is associated
with $Y$~\cite{kraemer01AmJPsychiatry}, a definition examined as a
nondiscrimination property by Skeem and
Lowenkamp~\cite{skeem16criminoloy}.
This definition will avoid the aforementioned issue, at least for alleles that are more strongly associated with a genetic condition than with race.

\subsection{Two-Side Correlation}
\label{sec:two-sided}

Datta~et~al.\@ use a special form of correlation for defining \emph{proxy discrimination}~\cite{datta17arxiv} (and \emph{use privacy}~\cite{datta17ccs}).
Their definition of \emph{perfect proxy}~\cite[Def.\,2]{datta17arxiv}, in our notion and giving their notion of proxy its own name, follows:
A variable $P$ is a \emph{proxy by (perfect) two-sided correlation} for a variable $T$ if and only if
there exist functions $f \colon \mathcal{P} \to \mathcal{T}$ and $g \colon \mathcal{T} \to \mathcal{P}$, such that $\Pr[T {=} f(P)] = 1$ and $\Pr[g(T) {=} P] = 1$.  
They later present a relaxed, quantitative form of proxy~\cite[Def.\,7]{datta17arxiv}.

Even in its quantitatively relaxed form, the two-sided nature of this
definition adopts a stricter view of \emph{proxy}, one in which the target
and proxy must each predict the other.
For example, %
assuming sex-segregated locker rooms,
the locker room used is a two-sided correlation proxy $P$ for the target $T$ of sex.
Similarly, the locker $L$ used reveals the user's
locker room and, thus, sex.
However, $L$ is not very predictable from sex $T$ since
each locker room has many lockers.
The used locker $L$ not being a two-sided correlation proxy for the target of sex
may seem counterintuitive since the point of proxies is to
predict the target, not the other way around.
In Datta~et~al.'s view (personal communication), it is more precise to
say that the variable $L$ can yield a proxy by using it to compute the
locker room used, which is a proxy for $T$.

Despite being more strict, this notion is similar to the associative notion above.
It also depends upon what population is used and can be made into a form conditional upon the other features $\vec{X}'$.
It also remains fairly broad.
Note that above example shows that, under this notion, the locker room used is a proxy for sex even when the classification task is predicting which building exit a gym user would like to be picked up at.
As with the example involving sickle cell anaemia, the connection between the putative proxy $P$ and the predicted class $Y$ is much stronger than putative target $T$, making it odd to see proxy discrimination here.
(That the sex segregation may be transphobic is a different problem.)

\subsection{Causal Descendants}
\label{sec:causal-paths}

Kilbertus et al.\@ define a \emph{proxy} to be ``nothing more than a descendant of [the target $T$] in the causal graph that we choose to label as a proxy''~\cite[p.\,4]{kilbertus17neurips}.
This differs from the aforementioned notions of proxy by looking at a causal relationship instead of at an associative one and by allowing for choice in which of them are so labeled.
Ignoring this normative choice in labeling, this notion of proxy is similar to their notion of potential proxy discrimination, which I discuss in Section~\ref{sec:use}.
Both notions are weak since they only require that being a causal descendant merely means that causation hasn't been ruled out, not that it exists (\S\ref{sec:background}, last paragraph).

The intuitions that Kilbertus et al.\@ have leading them to believe that a proxy $P$ should be a descendant of its target $T$ seem to be at odds with the intuitions of others that a proxy's capacity is determined by its ability to predict $T$.
For example, consider the target $T$ of \emph{will not have expensive healthcare costs}, used by an employer.
A reasonable proxy $P$ could be \emph{willing to gather carts}, as used by Walmart~\cite{greenhouse05nyt}.
However, neither causes the other, with both having the common cause of \emph{being in good health}.

\section{Proxy Use}
\label{sec:use}

While capacity provides a minimal conception of a proxy, the model using such a proxy provides a minimal conception of proxy discrimination. 
I found four different ways of making proxy use precise, each 
employing a different notion of causation (Table~\ref{tbl:proxy-use}).

\begin{table*}
\caption{Summary of notions of proxy use.  ``Gen.'' stands for ``generalization''.}
\label{tbl:proxy-use}
\footnotesize
\renewcommand{\arraystretch}{1.2} %
\begin{tabular}{@{}llr@{ }c@{ }ll@{}}
\toprule
Our name & From & \multicolumn{3}{l}{Holds when} & For any\\
\midrule
Potential use & \cite[Def.\,2]{kilbertus17neurips} & \multicolumn{4}{l}{$m$ gets $p$ or a causal descendant of $p$ as input}\\
Deterministic influence use & Gen.\@ of \cite[Def.\,3]{datta17arxiv} & $m(\vec{x}', p)(r)$ & $\neq$ & $m(\vec{x}', p')(r)$ & $\vec{x}, r, p, p'$\\
Probabilistic influence use & Gen.\@ of \cite[Def.\,3]{datta17arxiv} & $\Pr[\hat{Y}{=}\hat{y} \mid \mathrm{do}(P{=}p, \vec{X}'{=}\vec{x}')]$ & $\neq$ & $\Pr[\hat{Y}{=}\hat{y} \mid \mathrm{do}(P{=}p', \vec{X}'{=}\vec{x}')]$ & $\hat{y}, \vec{x}', p, p'$\\
Total effect use & \cite[Def.\,3]{kilbertus17neurips} & $\Pr[\hat{Y}{=}\hat{y} \mid \mathrm{do}(P{=}p)]$ & $\neq$ & $\Pr[\hat{Y}{=}\hat{y} \mid \mathrm{do}(P{=}p')]$ & $\hat{y}, p, p'$\\
\bottomrule
\end{tabular}
\end{table*}

Each of them might be too sensitive for practical use.
Some ML algorithms will use any feature with even a small association with $Y$ conditional upon other features.
Ensemble methods may use multiple training data samples or multiple sets of other features while looking for associations, providing multiple chances to accuse a single potential proxy and leading to something similar to the proxy fishing discussed in Section~\ref{sec:correlation}.
For such algorithms, a low-level use may occur for all the available features.
An auditor may need to prioritize examining them by the size of the use, which could be measured with any of the ways of measuring effect sizes or the influence of features in models~\cite[e.g.,][]{datta16sp}.

\subsection{Influence}
\label{sec:influence}

Datta et al.\@ define \emph{influence} for a deterministic model $m$.~\cite[Def.\,3]{datta17arxiv}.
We can represent such a model as a function from the features $\vec{X}$ that it uses to reach its prediction $\hat{Y}$, that is, $\hat{Y} := m(X_1,X_2,\ldots,X_n)$ where $\langle X_1,X_2,\ldots,X_n\rangle = \vec{X}$ are the features used as arguments to the model viewed as a function.
Let the $i$th feature $X_i$ be the proxy $P$.
They say that $P$ has \emph{influence} on $\hat{Y}$ iff there exists values $x_j$ for $X_j$ for all $j \neq i$ and a pair of values $p_i$ and $p'_i$ for $P$ such that 
\begin{align}
m(x_1, x_2, \ldots, x_{i-1}, p, x_{i+1},\ldots,x_n) &\neq m(x_1,x_2,\ldots,x_{i-1},p',x_{i+1},\ldots,x_n)
\label{eqn:influence}
\end{align}

This definition can be generalized to randomized models in two different ways.
The first makes the source of randomness explicit as a random variable $R$ and treats it as another input to $m$ that is independent from the others.
The comparison~\eqref{eqn:influence} would hold the value of $R$ constant on both sides of the inequality, just like any input other than $P$, which effectively removes the randomization.
This generalization is related to the negation of \emph{deterministic causal irrelevance}~\cite[Def.\,13]{galles97ai}, a standard notion found in research on causal modeling.

The second generalization compares the distributions of outputs that $m$ randomly produces.
To make this characterization precise, let $\vec{X}'$ be the features $\vec{X}$ other than $X_i = P$.
Our comparison now checks whether there exists $\hat{y}$, $\vec{x}'$, $p$, and $p'$ such that
$\Pr[\hat{Y}{=}\hat{y} \mid P{=}p, \vec{X}'{=}\vec{x}'] \neq \Pr[\hat{Y}{=}\hat{y} \mid P{=}p', \vec{X}'{=}\vec{x}']$,
where the probabilities are over the randomization internally used by $m$.
Since all the inputs are conditioned upon, there's no other sources of randomization, and
it is equivalent to checking whether
$\Pr[\hat{Y}{=}\hat{y} \mid \mathrm{do}(P{=}p, \vec{X}'{=}\vec{x}')] \neq \Pr[\hat{Y}{=}\hat{y} \mid \mathrm{do}(P{=}p', \vec{X}'{=}\vec{x}')]$~\cite[Property~1, p.\,24]{pearl09book}.
This generalization is related to the negation of \emph{probabilistic causal irrelevance}~\cite[Def.\,7]{galles97ai}, to the \emph{controlled direct effect}~\cite[p.\,133]{pearl09stat-surveys}, and to \emph{probabilistic interference}~\cite[Def.\,2]{tschantz15csf}, a property used in computer security research.

This probabilistic notion of influence implies the deterministic notion, but not the other way around.
For example, suppose the output of $m$ is equal to $P \mathop{\mathrm{xor}} R$ where both variables are binary and $R$ is uniformly random.
That $R$ could be either $0$ or $1$ would hide the value of $P$ and produce identical uniformly random distributions for the probabilistic version, but would show influence for the deterministic version since it compares the outputs under one fixed resolution of the randomness of $R$ at a time.
Intuitively, probabilistic influence does not occur if the only effect of $P$ is to switch which individuals within a group of identical looking (as far as $\vec{X}$ is concerned) individuals get which classification without changing how common each classification is for that group, suggesting that it aligns more with group fairness whereas the deterministic version aligns more with individual fairness.
(This is somewhat related to the difference between Individual Treatment Effects and Conditional Average Treatment Effects~\cite{vegetabile21arxiv}.)

\subsection{Total Effects}
\label{sec:total-effects}

Since influence does not always imply correlation, we might wish to demand that $P$ and $\hat{Y}$ be correlated before concluding that $m$ uses $P$ to determine $\hat{Y}$ in a meaningful sense.
For example, suppose that $\hat{Y} = m(P, X_1) = P + X_1$.
Comparing $m(0, 0) = 0$ to $m(1, 0) = 1$ shows that $\hat{Y}$ is caused by $P$ under $m$, for causation defined in terms of either form of influence.
However, suppose that in the population on which $m$ is to be used, $P$ has the effect of setting the value of $X_1$ to $-P$.
Then, $P$ and $X_1$ cancel out in the computation of $\hat{Y}$, and $P$ will not be correlated to $\hat{Y}$ in that population.
Viewing such canceling out as absolving, Kilbertus et al.~\cite[Def.\,3]{kilbertus17neurips} look at whether 
$\Pr[\hat{Y}{=}\hat{y} \mid \mathrm{do}(P{=}p)] \neq \Pr[\hat{Y}{=}\hat{y} \mid \mathrm{do}(P{=}p')]$.
This definition of proxy use is related to the \emph{total effect} of $P$ on $\hat{Y}$~\cite[p.\,132]{pearl09stat-surveys}.
By not fixing the values of the other features $\vec{X}'$, it allows those of them that depend upon $P$ to vary with it, allowing such canceling out.

This notion is neither broader nor more narrow than the influence notions.
We already saw a case where it does not hold despite there being influence.
For an example in the opposite direction, consider a case where $m(P, X_1) := X_1$ and $X_1 := P$.
By overwriting the value of $X_1$ with a causal intervention, the influence definitions hide $m$'s indirect use of $P$.
Total effects differ by looking at not just what happens within $m$ but also before it as features may affect one another.

\subsection{Causal Descendants}
\label{sec:use-causal-descendants}

Kilbertus et al.\@ also consider a very broad notion of use.
They consider it ``potential proxy discrimination'' if $\hat{Y}$ is a causal descendant of $P$ in a graphical model~\cite[Def.\,2]{kilbertus17neurips}.
Recall (\S\ref{sec:background}, last paragraph), that this does not imply that changing $P$ will result in a change in $\hat{Y}$.
Rather, it asks whether $P$ or a descendant of $P$ is provided as an input to $m$. 
For simplicity, I have been treating $P$ as an input to $m$, nothing much changes if we allow considering the causes of inputs to potentially be proxies as well, as Kilbertus et al.\@ do.
(This is another difference with Datta et al.\@ who instead considers proxies constructed within $m$ from multiple inputs~\cite[Def.\,4]{datta17arxiv}.  
I focus on just the inputs of $m$ to focus on $P$'s relationships to other variables and not where it resides.)

\section{Target Unused}
\label{sec:target-unused}

The third element of proxy discrimination that I consider is whether the model $m$ uses the supposed target $T$.
Many discussions of proxies start with the presupposition that the target is unused~\cite[e.g.,][]{oneil16hbr}, making this at least an implicit pre-condition for any feature to be a proxy of $T$.
I do not view this element as strictly required for $P$ to be a proxy for $T$ since some discussions of proxies do not mention whether $T$ is used~\cite[e.g.,][p.\,16]{kirkpatrick16cacm}\qnote{It is not only the criminal justice system that is using such algorithmic assessments. Algorithms also are used to serve up job listings or credit offers that can be viewed as inadvertently biased, as they sometimes utilize end-user characteristics like household income and postal codes that can be proxies for race, given the correlation between ethnicity, household income, and geographic settling patterns.}.
Furthermore, as mentioned in Section~\ref{sec:overview}, a model can simultaneously use $T$ and use $P$ because it is correlate with $T$, which sounds like a reasonably strong case for proxy discrimination.
We can make using the target precise with the notions of use discussed above.

\section{Bottom Line Impact}
\label{sec:impact}

Since $P$ having the capacity to target $T$ is a low bar to declare proxy discrimination, even if $P$ is used and $T$ unused, we need additional elements that focus our attention on the most problematic cases.
Requiring a bottom line impact is one such restriction.
A bottom line impact indicates that the proxy discrimination was successful in the sense of there being an association between $\hat{Y}$ and $T$.

To further motivate looking at the bottom line impact, note that proxies can be related to \emph{disparate impact} by viewing 
the target $T$ as a protected attribute (e.g., race),
the estimate $\hat{Y}$ as some outcome (e.g., getting fired or not hired),
the model $m$ as how the company decides who gets which outcomes,
and the proxy $P$ as a feature used by $m$ to reach such decisions,
where the use of $P$ by $m$ would be the business practice.
When $P$ is sufficiently associated with $T$, the use of $P$ by $m$ to determine $\hat{Y}$ would, prima facie, have a disparate impact upon the protected group identified $T$, provided the correlation is in the adverse direction.
(Additional conditions must hold to win a disparate impact case, such as there not being a \emph{business necessity} to use $P$~\cite[e.g.,][]{barocas16cal-l-rev}.)

The U.S.\@ Equal Employment Opportunity Commission (EEOC), as a matter of policy, maintains that it will only bring cases when a disparate impact also appears in the \emph{bottom line} comparison of $\hat{Y}$ to $T$~\citetext{\citealp{eeoc78cfr}, \citealp{eeoc79fr}\qnote{Adverse impact is determined first for the overall selection process for each job. [\ldots]  If there is no adverse impact for the overall selection process, in most circumstances there is no obligation under the Guidelines to investigate adverse impact for the components, or to validate the selection procedures used for that job.}}.
(See~Fig.~\ref{fig:ov-eeoc}.  The EEOC's policy is not binding upon individual private plaintiffs who can show prima facie disparate impact without showing bottom line impact~\cite{brennan82scotus,tschiemer83ark-lr}.)
Since $\hat{Y}$ might depend upon factors other than $P$, this extra third condition on top of use and capacity is not trivial and allows the EEOC to focus on cases where the targeted group suffers an adverse outcome disproportionately.
Similarly, one might wish to focus on just the proxies $P$ whose target $T$ is correlated with the estimate $\hat{Y}$.

\section{Consequential}
\label{sec:consequential}

One might wish to further focus on cases where there's not just a bottom line impact but the proxy use induces it.
In this case, the proxy use is particularly consequential, the first second-order element of proxy discrimination that I will detail.
I can make this intuition more precise with any of the causal notions discussed in Section~\ref{sec:use}.
Below, we will take a closer look at total effects.
We will also look at using intentions to make this inducement more precise.

\subsection{Total Effects}

We could look at the total effect of using $P$ on the correlation between $T$ and $\hat{Y}$.
The meaning of an use having an effect on a correlation can be intuitively explained in terms of an experiment.
Whereas experiments typically work with a sample of individuals (e.g., humans or mice), our experiment will be second order in the sense of working with a sample of data sets of individuals. 
The experiment randomly assigns each data set, as a whole, to either an experimental or a control condition.
For each data set $j$ in the control group, an estimator $\hat{Y}_j$ over individuals is computed for the data set $j$, producing a classified data set.
Each classified data set resulting from classifying the control group is checked for a correlation between $\hat{Y}_j$ and $T$.
For each data set $j$ in the experimental group, an estimator $\hat{Y}_j$ is computed that also operates over each individual in the data set but with the value of $P$ eliminated, producing a data set classified in a modified manner, which is checked for a correlation between $\hat{Y}_j$ and $T$.
If the number of correlations is statistically significantly lower for the experiential group than the control group, then the proxy use causes correlations.

Just as there's more than one way to administer a drug during a clinical trial, there's more than one way to eliminate $P$ from the data provided to the model to produce the estimations $\hat{Y}_j$ in the experimental group.
The most obvious is to delete the value of $P$ from each data point, but this may force the model into an error condition if it depends upon a value being available.
In some cases, the model might work properly if the value of $P$ is instead replaced by a null value meaning that the value is missing.
If not, then the value for each individual could be replaced by the average value of $P$ over all individuals. 
Each of these methods lead to different operationalizations of using $P$ and could lead to different results.

Note that the above experiment does not retrain the model $m$ producing the $\hat{Y}_j$ for the experiential group.
This is because we focused on whether a fixed model $m$'s use of $P$ caused an association.
If we were instead interested in whether the ML algorithm $a$'s use of $P$ to create a model caused the association, we would instead retrain the model for each data set, with the value of $P$ removed for those in the experimental group.

\subsection{Intentional}

We could instead ask whether the intention behind the use of the proxy is to produce a bottom line impact.
Hellman appears to have such intentions in mind when defining ``proxy discrimination'' as involving selecting one group ``\emph{in order to} reach another group''~\cite[pp.\,317--8, emphasis in original]{hellman98CaLR}\qnote[emphasis in original]{This is a case of proxy discrimination because the firm selects the group identified by the classification, those in the top 10\% of the class, \emph{in order to} reach another group: those who will be effective lawyers.  I call this form of discrimination ``proxy discrimination'' because the firm uses one identifying characteristic as a proxy for another.}.
While her work only touches on proxies in an informal manner, it appears that 
Lipton~et~al.\@ also had such intentions in mind~\cite[pp.\,2\,\&\,9]{lipton18arxiv}.

The intention could be held by the decisionmaker, ML algorithm, and/or model.
While the first comes to mind most readily,
there is no philosophical bar to ascribing intentions to non-living actors~\cite{dennett87book} and
recent research has examined assigning intentions to algorithms~\cite{ashton20jurisin,ashton21arxiv}.
Numerous formalisms exist for capturing intentions and related concepts that could be used in the place of intentions, such as desires or purposes~\cite[e.g.,][]{bratman87book,roy08phd,tschantz12sp}.

This alternative version of inducement is neither broader nor narrower than a causal version:
it's possible that the decisionmaker intends to have a bottom line impact but fails to cause one, and it's possible that using the proxy accidentally caused a bottom line impact.
Given that authors distinguish between intentional and unintentional proxy discrimination, intent seems be viewed as an aggravating factor, not a requirement, for proxy discrimination~\citetext{e.g., 
\citealp[p.\,675]{barocas16cal-l-rev}\qnote{Even in situations where data miners are extremely careful, they can still effect discriminatory results with models that, quite unintentionally, pick out proxy variables for protected classes.}; 
\citealp[p.\,16]{kirkpatrick16cacm}\qnote%
{It is not only the criminal justice system that is using such algorithmic assessments. Algorithms also are used to serve up job listings or credit offers that can be viewed as inadvertently biased, as they sometimes utilize end-user characteristics like household income and postal codes that can be proxies for race, given the correlation between ethnicity, household income, and geographic settling patterns.}}.

Such intentions have to do not with the relationship between $P$ and $T$ themselves, but rather with why $P$ is used by the model.
Thus, this concept bleeds into the causes of the proxy use, which I cover in the next two sections.

\section{Capacity Induced Proxy Use}
\label{sec:mot}

The next relationship I consider is whether the proxy $P$'s capacity to target $T$ induced the proxy's use, which I take to mean that the proxy is acting as a proxy.
Prince and Schwarcz define \emph{proxy discrimination} in terms of the proxy's capacity being the reason that the proxy is useful to the decisionmaker~\citetext{\citealp{prince20iowa-lr}, p.\,1261\qnote[footnote omitted]{In particular, proxy discrimination requires that the usefulness to the discriminator of a facially neutral practice derives, at least in part, from the very fact that it produces a disparate impact.} \& p.\,1270\qnote{This would constitute proxy discrimination, because it would (1) disparately impact members of a protected group, and (2) prove useful to the firm for precisely this reason.}}.
If we take usefulness to be the reason that the model uses the proxy, we can view their definition as an instance of capacity induced proxy use where the inducement is an indirect form of causation flowing from the capacity to the usefulness to the use.

We can make this flow more precise by examining how usefulness can lead to use.
Prince and Schwarcz identify two ways.
In the first way, the decisionmaker, upon finding $P$'s capacity to target $T$, forces the model $m$ to use $P$ by either hand crafting $m$ or selecting an ML algorithm that will have $m$ use $P$ regardless of whether doing so furthers the algorithm's central goal of accurately estimating $Y$.
This is what Prince and Schwarcz call ``intentional proxy discrimination'', which they consider to be a form of disparate treatment~\cite[p.\,1269]{prince20iowa-lr}\qnote{Intentional proxy discrimination clearly violates most anti-discrimination laws because it constitutes disparate treatment.}.
Since standard ML algorithms focus only on selecting features that predict $Y$, if $P$ does not do so, then achieving such intentional proxy discrimination with ML requires a nonstandard algorithm, at least if the proxy use being induced is to be meaningfully large and reliable (recall the second paragraph of \S\ref{sec:use}).

In the second way, the decisionmaker uses a standard ML algorithm that decides to have $m$ use $P$ because it increases its accuracy, ``making discrimination `rational'\,''~\cite[p.\,1262]{prince20iowa-lr}.
Prince and Schwarcz call this way ``unintentional proxy discrimination'', which they consider to be a form of disparate impact~\cite[p.\,1272]{prince20iowa-lr}\qnote{unintentional proxy discrimination represents one specific type of disparate impact claim.}.
They discuss various manners in which the association between $P$ and $T$ can cause an association between $P$ and $Y$, which causes a standard ML algorithm to use $P$~\cite[pp.\,1277--81]{prince20iowa-lr}. 
Intuitively, these all boil down to $T$ fully mediating the association between $P$ and $Y$ so that conditioning upon $T$ would remove the association between $P$ and $Y$, but conditioning on $P$ would not remove it from between $T$ and $Y$~\cite[p.\,1262]{prince20iowa-lr}\qnote[footnote omitted]{[\ldots] data on applicants’ membership in the Facebook group would likely cease to be predictive of claims in a model that controlled for applicants’ genetic predispositions to cancer.}.
From this perspective, the associations between $P$ and $T$ and between $T$ and $Y$ are more fundamental than that between $P$ and $T$, which can be taken to mean that the first two induce the third.
This provides an observational test for capacity induced use.

A more experimental approach comes to mind if one wishes to use a more causal form of inducement.
An auditor could provide the ML algorithm with various data sets, some where $P$ has a capacity to proxy $T$ and some lacking this capacity.
If the models produced from data with the capacity uses $P$ and those produced without it do not, then, intuitively, the capacity causes the use.
Recall that standard ML algorithms will check whether $P$ is associated with $Y$, meaning that the experiment's results will largely depend upon whether breaking the association between $P$ and $T$ also breaks it between $P$ and $Y$.
We would expect this to happen if we are in a setting where $T$ mediates the association between $P$ and $Y$, showing a link between this experimental test and the observational tests discussed above.

A shortcoming of this approach is that it fails to check for whether the decisionmaker selected the algorithm to exploit the capacity, making it more appropriate for testing for unintentional proxy discrimination than intentional proxy discrimination.
Another challenge is that how the capacity is removed can determine whether or not such an experiment will find causation, but there is 
no obvious canonical way to eliminate a capacity to proxy from data sets.
In fact, as discussed in Section~\ref{sec:correlation}, there's many different ways of measuring the capacity even when limiting ourselves to correlation-like capacities.
Further work is needed to make this second-order notion of causation precise, providing a mathematical definition of when a proxy's capacity causes its use.

\section{Proxy Substituting for Target}
\label{sec:substituting}

I say that a proxy is \emph{substituting} for the target if the model's lack of using the target $T$ induces it to use the proxy $P$.
An obvious reason for such substitution is that the model wants to use $T$ and upon being denied access to it uses $P$ due its correlation with $T$, that is, capacity induced use. 
In such cases, substitution and capacity induced use work hand in hand as the lack of using $T$ and its correlation with $P$ jointly cause $P$'s use.
One approach to understanding such substitution is the study of omitted variable bias, covered in more detail below.

It's also possible to have substitution without capacity inducement and capacity inducement without substitution, as shown in the examples of Section~\ref{sec:overview}.
Indeed, ML algorithms can react arbitrarily to the presence or absence of the target $T$.
We can study such reactions using second-order experiments similar to those discussed in Section~\ref{sec:total-effects}, but where we look for whether a lack of use causes a different use, instead of whether a use causes a correlation.  
I leave making this mathematically precise to future work.

\subsection{Omitted Variable Bias}

Going back to at least an FTC report~\cite[p.\,61]{ftc07report},\qnote{The included variable thus may act in whole or in part as a statistical `proxy' for the excluded variables of race, ethnicity, or income. [Footnote:] The econometric term for this effect is `omitted variable bias.' [\ldots]} 
a series of papers has viewed proxies in terms of omitted variable bias~\cite{ftc07report,pope11am-econ-j,morris17j-empirical-legal-studies}.
The basic idea is that if $T$ is associated with $Y$ but not in the set of features $\vec{X}$, then the ML algorithm might construct a proxy $P$ of $T$ for the model to use to recreate the usefulness of $T$ in predicting $Y$.
Under this view, to measure the degree to which $P$ proxies for $T$, an auditor compares how models constructed by the algorithm with and without access to $T$ differ in their use of $P$.
For each model, the auditor computes some measure of how much the model uses $P$.
For example, Morris~et~al.\@ used the size of $P$'s coefficients in the model $m_{\mathsf{w/} \,T}$ with access to $T$ and in the model $m_{\mathsf{w/o} \,T}$ without access to $T$~\cite{morris17j-empirical-legal-studies}.
Then, the auditor checks whether these usage measures are close in value.
If so, then not letting the algorithm use $T$ in the model $m_{\mathsf{w/o} \,T}$ has little effect on the model's usage of $P$, and the model $m_{\mathsf{w/o} \,T}$ uses $P$ for reasons other than proxying for $T$.
If the usage of $P$ deceases from the model $m_{\mathsf{w/o} \,T}$ without $T$ to the model $m_{\mathsf{w/} \,T}$ with $T$, then the algorithm has shown a preference for models to use $T$ over $P$, letting the auditor conclude that the use $P$ is proxy for $T$ in $m_{\mathsf{w/o} \,T}$.

This notion of proxy has some limitations.
The notion assumes that the algorithm will switch to using $T$ and away from using $P$ only when $T$ is more closely related than $P$ is to the true class labels $Y$. 
This assumption is reasonable since the feature with the tighter connection to $Y$ will typically be strictly more useful for predicting it.
However, the assumption can be violated either by ML algorithms that attempt to use multiple features simultaneously or when $T$ and $P$ are equally accurate for $Y$.
For example, if $Y := X_1 := X_2$, %
then, despite $X_1$ being closer to $Y$ in the chain of causation, $X_1$ and $X_2$ are equally useful for predicting $Y$.
Such \emph{multicollinearity} can cause ridge regression to assign the coefficient of $0.5$ to both $X_1$ and $X_2$ when offered both features and no others.
When run on just one of them, the regression will assign the coefficient of $1$ to it.
Thus, depending upon whether the auditor starts with $X_1$ and checks whether it is a proxy for $X_2$ or the other way around, the auditor will either find that $X_1$ is a proxy for $X_2$ or that $X_2$ is a proxy for $X_1$.
Domain knowledge is required to decide which is the case.
Worse, under lasso regression the coefficients will be highly unstable with the regression possibly assigning a coefficient of $0$ to either of them when offered both.

\section{When and Why is Proxy Discrimination Wrong or Illegal?}
\label{sec:norms}

There is general agreement that the intentional use of proxies to harm protected groups out of animus is wrong, and often illegal~\cite[e.g.,][p.\,456]{alexander97constitutional-commentary}.
The difficult case is unintentionally, or even unknowingly, using a proxy to achieve some acceptable ends.

The natural starting point for such proxy use is disparate impact, a form of illegal employment discrimination where a business practice adversely affects a protected group to a disproportionate degree.
It is legal if the practice is a \emph{business necessity}, a concept that has become watered down to include almost anything that furthers a business's goals~\cite[e.g.,][]{grover96geo-lr}.
Some authors consider proxy discrimination to be a particularly concerning form of disparate impact~\cite[e.g.,][]{prince20iowa-lr,barocas16cal-l-rev}.
Using machine learnt models, which often use proxies, appears to be such a business necessity, allowing those using proxies to largely avoid liability under disparate impact~\citetext{\citealp[p.\,709]{barocas16cal-l-rev}\qnote{Thus, there is good reason to believe that any or all of the data mining models predicated on legitimately job-related traits pass muster under the business necessity defense.};
\citealp[p.\,866]{kim17wm-mary-lr}\qnote[footnote omitted]{As a result, [data mining models] pose a different set of risks---risks that existing doctrine does not address well.  [¶]
As an example, disparate impact doctrine provides a defense if an employer can show that a test is “job related \ldots and consistent with business necessity.” In the case of workforce analytics, the data algorithm by definition relies on variables that are correlated in some sense with the job. So to ask whether the model is “job related” in the sense of “statistically correlated” is tautological.};
\citealp[p.\,1305]{prince20iowa-lr}\qnote[footnote omitted]{By contrast, disparate impact liability (as it is currently constructed) is simply not capable of effectively policing against proxy discrimination by AIs. The central problem is that firms using AIs that proxy discriminate will typically have little problem showing that this practice is consistent with business necessity and in rebuffing any attempt to show the availability of a less discriminatory alternative.}%
}.
Westreich and Grimmelmann would not let employers off so easily and suggest that the burden of showing a business necessity should require that the employer can explain why its model is not merely using some factors as a proxy for race~\cite{westreich17cal-lr-online}.\qnote[emphasis in original]{We believe that where a plaintiff has identified a disparate impact, the defendant’s burden to show a business necessity requires it to show not just that its model’s scores are not just \emph{correlated} with job performance but \emph{explain} it.}
If they fail, then they should be assumed to be effectively using a protected attribute, which is illegal, they argue.

Authors present numerous other arguments for why such proxy discriminatory is, is not, or might be wrong.
Both Prince and Schwarcz~\cite[p.\,1283]{prince20iowa-lr}\qnote{Ultimately, the argument is straight-forward: Laws that prohibit discrimination based on directly predictive traits are normatively grounded in the goal of preventing specific outcomes for members of protected groups. Unlike some anti-discrimination settings, the questions of how or why bad outcomes are experienced by protected groups are secondary, if relevant at all, in these domains. Because proxy discrimination by AIs tends to produce the very same outcomes that would result in the absence of legal restrictions on discrimination based on directly predictive traits, it represents a substantial threat to the normative underpinnings of these anti-discrimination regimes.}
and Feldman et al.~\cite[p.\,6]{feldman15kdd}\qnote{More importantly, from a functional perspective, it does not matter whether Alice uses $X$ explicitly or uses proxy attributes $Y$ that have the same effect: this is the core message from the Griggs case that introduced the doctrine of disparate impact.}
point out that proxy discrimination can lead to the exact same outcomes as an anti-discrimination provision was intended to prevent, which suggests a focus on the bottom line impact and whether the proxy use was consequential.
These arguments seem to view proxy discrimination as an example of why disparate impact must be reformed, rather than as a form of discrimination that warrants separate attention.
With more focus on process than outcomes, both Westreich \& Grimmelmann~\cite[p.\,176]{westreich17cal-lr-online}\qnote{In our view, Title VII does not permit an employer to do indirectly what it could not do directly.}
and
Morris~et~al.~\cite[p.\,7, fn.\,7]{morris17j-empirical-legal-studies}\qnote{An important regulatory principle is that insurers cannot be allowed to do indirectly what they are prohibited from doing directly.}
emphasize 
that employers should not be allowed to do indirectly what is prohibited to do directly and view proxy use as indirectly using a prohibited attribute, which suggests more focus on whether the proxy's capacity induced the use.

Alexander and Cole see a subtle difference in the moral consequences of proxy discrimination, making it not merely achieving indirectly what cannot be done directly~\cite{alexander97constitutional-commentary}.
They believe that ``rational'' discrimination is illegal to catch the irrational sort
and to prevent the reinforcement of negative attitudes that come from the ``\emph{overt} use'' of a sensitive attribute~\cite[p.\,457, emphasis in original]{alexander97constitutional-commentary}.
They point out that proxy discrimination is often rational (e.g., the unintentional sort; see \S\ref{sec:mot}) and too covert to reinforce negative attitudes.
While they emphasize that such covert proxy discrimination is unstable -- it's only a matter of time before someone realizes what has happened -- they hesitate to condemn it while it exists~\cite[p.\,462]{alexander97constitutional-commentary}.\qnote{We can feel quite confident that School E is not acting out of bias, pronounced or subtle. Its ignorance establishes innocence. And to the extent that untoward social effects are mainly the product of social awareness of the link between race and goal, that factor is also missing. But placing a premium on ignorance is an uncomfortable constitutional position. Moreover, once School E becomes aware of why its approach works, the morality of its continuing the approach seems virtually indistinguishable from the morality of School C's adoption of the policy in the first place.}
Furthermore, they leave open the possibility that such proxy use might remain acceptable after its discovery~\cite[p.\,463]{alexander97constitutional-commentary}.\qnote[footnote omitted]{When that correlation is direct, so that the non-racial discrimination serves permissible ends in direct relation to its correlation with racial discrimination -- the case of School C's regional policy -- the two principles seem to collide. [¶] It is time, therefore, to rethink the issue of rational discrimination. [\ldots] demanding that government ignore race and sex but also act rationally will be demanding the impossible.}
(Cf.\@ Westreich and Grimmelmann who may object that this discourages employers from understanding their models~\cite[p.\,176]{westreich17cal-lr-online}.\qnote{We do not want to deter employers from examining closely the models and algorithms they use. A test that turned only on the employer’s knowledge of how its model functions would discourage employers from looking too closely at models that superficially seemed to work.})
Alexander and Cole's views suggest that auditors should focus on finding what Prince and Schwarcz called ``intentional proxy discrimination'' (\S\ref{sec:mot}).

\section{Conclusion}
\label{sec:conclusion}

I find proxy discrimination to be characterized by numerous relationships between variables and between relationships, without any subset of them being clearly necessary and sufficient (\S\ref{sec:overview}).
My examination of normative justifications for concern over proxy discrimination yields varying suggestions for which relationships to focus on (\S\ref{sec:norms}).
Furthermore, each of these relationships can be made precise in numerous manners (\S\S\ref{sec:capacity}--\ref{sec:substituting}, esp.\,\S\ref{sec:capacity}), with no clearly canonical ones.

In general, I believe it becomes increasingly justified to invoke \emph{proxy discrimination} as the importance and sensitivity of the attributes $Y$, $\hat{Y}$, and $T$ increases, as the number of elemental relationships shown (Tbl.~\ref{tbl:elements}) that hold increases, and as the senses in which these relationships hold become stronger.
However, I do not find any firm cutoffs and am not in the position to impose them.
Since many of these requirements will often unintentionally be met, at least weakly, it may be difficult to avoid using features that are not at least weakly proxies.

While there may be some uses for detecting or suppressing well-crafted subsets of the most concerning forms of proxy discrimination, as they will also be particularly concerning cases of impact disparity, research attempting to precisely define, catch, and eliminate proxy discrimination per se might be misusing the concept.
Such research will have to either adopt an artificially narrow conception of proxy discrimination, which would be better called something else, or disallow almost all classification.
I don't believe this to be a flaw with the concept of proxy discrimination since it largely arose as a lament about how unavoidable impact disparities are under ML.
\emph{Proxy discrimination} remains a useful term for invoking a cluster of concerns, perhaps an essentially contested one, even if it's not useful, on its own, for categorizing features or models as those to be prohibited.
\begin{acks}
I thank Anupam Datta, Piotr Mardziel, and Shayak Sen for discussions about their work.
I thank Margarita Boyarskaya, Solon Barocas, and Hanna Wallach for sharing their ongoing work with me.
I gratefully acknowledge funding support from the 
\grantsponsor{NSF}{National Science Foundation}{https://www.nsf.gov/}
(Grant \grantnum{NSF}{1704985})
and the
\grantsponsor{DARPA}{Defense Advanced Research Projects Agency}{https://www.darpa.mil/}
(Grant \grantnum{DARPA}{FA8750-15-2-0277}).
The views and conclusions contained herein are those of the author and should not be interpreted as necessarily representing the official policies or endorsements, either expressed or implied, of the Defense Advanced Research Projects Agency (DARPA), the Air Force Research Laboratory, NSF, or the U.S.\@ Government.
\end{acks}
\bibliographystyle{ACM-Reference-Format}

\end{document}